# An attention mechanism based convolutional network for satellite precipitation downscaling over China

Yinghong Jing[a], Liupeng Lin[a], Xinghua Li[b], Tongwen Li[c], Huanfeng Shen[a]*

[a] School of Resource and Environmental Sciences, Wuhan University, Wuhan 430079, China
[b] School of Remote Sensing and Information Engineering, Wuhan University, Wuhan 430079, China
[c] School of Geospatial Engineering and Science, Sun Yat-sen University, Zhuhai 519082, China
* **Corresponding author:** Huanfeng Shen (shenhf@whu.edu.cn)

**Abstract**
Precipitation is a key part of hydrological circulation and is a sensitive indicator of climate change. The Integrated Multi-satellitE Retrievals for the Global Precipitation Measurement (GPM) mission (IMERG) datasets are widely used for global and regional precipitation investigations. However, their local application is limited by the relatively coarse spatial resolution. Therefore, in this paper, an attention mechanism based convolutional network (AMCN) is proposed to downscale GPM IMERG monthly precipitation data from 0.1° to 0.01°. The proposed method is an end-to-end network, which consists of a global cross-attention module, a multi-factor cross-attention module, and a residual convolutional module, comprehensively considering the potential relationships between precipitation and complicated surface characteristics. In addition, a degradation loss function based on low-resolution precipitation is designed to physically constrain the network training, to improve the robustness of the proposed network under different time and scale variations. The experiments demonstrate that the proposed network significantly outperforms three baseline methods. Compared with in-situ measurements, the root-mean-square error is decreased by 5.41–19.82 mm/month in the real-data experiment. Finally, a geographic difference analysis method is introduced to further improve the downscaled results by incorporating in-situ measurements for high-quality and fine-scale precipitation estimation.

## 1. Introduction

Precipitation is an essential factor driving the hydrological cycle, which also modulates the surface energy balance, is closely connected with climate change, and affects human activities (Lincoln, 2008; Madakumbura et al., 2021). Despite the compelling evidence for the impact of global warming on large-scale precipitation patterns, the investigation of regional patterns is restricted by the observational roughness and modeling uncertainties (Sarojini et al., 2016). Ground measurement networks are distributed sparsely and unevenly across different regions, and are also varied in quality. Precipitation modeling data typically have a relatively low spatial resolution and an uncertain accuracy. These limitations have resulted in satellite observations being considered as crucial materials for regional precipitation exploration (Sorooshian et al., 2011). The Global Precipitation Measurement (GPM) mission integrates infrared, microwave, and in-situ measurements, partly overcoming the shortcomings of an individual sensor. With their global coverage, high temporal resolution, and reliable accuracy, GPM precipitation datasets are increasingly being used in geoscience research. However, the relatively coarse spatial resolution means that these datasets cannot easily meet the diverse requirements of various local Earth science applications, resulting in the requirement for the downscaling of satellite precipitation data.

Various downscaling algorithms have been proposed over the past decades, including dynamic methods, statistical models, and machine learning based algorithms. The dynamic downscaling methods simulate complex physical processes through hydrological models, regional climate models, or land-atmosphere coupling models, integrating multi-source data from satellite images, model outputs, and/or in-situ measurements. High-resolution (HR) images are then sequentially generated using data assimilation techniques such as the ensemble Kalman filter to provide the initial field for the dynamic model. As a result, fine-scale precipitation data with physical consistency are constantly being updated (Wang et al., 2021). For example, Boussetta et al. (2008) innovatively constructed a coupled land-atmosphere model for better precipitation projection through assimilating Tropical Rainfall Measuring Mission (TRMM) Microwave Imager (TMI) data. Such methods take into account both the model and measurement uncertainties and have strong physical significance. They are thus widely used for the downscaling of global climate models. However, the disadvantages are the challenging parameter acquisition, complex internal mechanisms, and high computational requirements. In contrast, the statistical downscaling methods, assisted by various multi-scale environmental contextual factors, statistically model the linear or nonlinear relationships between the cofactors and the target variable at a coarse scale, and then estimate the target variable with a scale-invariant relationship model at a fine scale. Due to their simplicity and efficiency, these methods are now dominant in downscaling research (Chaudhuri and Srivastava, 2017; Ma et al., 2020; Chen et al., 2020). In recent decades, statistical models have developed from multiple linear regression models (MLR, Jia et al., 2011) to complicated nonlinear models (Wang et al., 2019). For example, Immerzeel et al. (2009) constructed an exponential regression (ER) model between precipitation and the normalized difference vegetation index (NDVI) for downscaling. Geographically weighted regression models (GWR, Xu et al., 2015; Chen et al., 2015) and their advanced models (Zeng et al., 2022) have also been widely applied in precipitation downscaling research. In addition, Zhang et al. (2018a) demonstrated that the nonlinear quadratic parabolic profile model outperforms the ER, MLR, and GWR models, because the relationship between precipitation and predictors is nonlinear. Tan et al. (2022) introduced the multivariate adaptive regression spline (MARS) model for precipitation downscaling with nine predictors, including vegetation, temperature, water, and topography. However, such methods cannot easily explain the complicated nonlinear relationships among the various factors, and are limited by the resolution and the quality of the predictors.

With the explosive growth of "big Earth data", increasing numbers of covariables are being adopted in downscaling research. However, the conventional statistical models cannot easily represent the complicated features of high-dimensional data. Machine learning methods were introduced into downscaling research due to their superior nonlinear fitting ability and have improved the downscaling performance (Ghorbanpour et al., 2021). For example, the random forest (RF) model has been used to explore the nonlinear relationships of precipitation and predictors, including vegetation cover, surface temperature, and geographic/topographic factors (Zhang et al., 2019; Chen et al., 2021; Zhao, 2021; Yan et al., 2021). Shen and Yong (2021) discovered that the gradient boosting decision tree and RF models are generally superior to support vector machine (SVM). Jing et al. (2016) demonstrated that the RF model generally performs better than the other common machine learning algorithms used for downscaling, followed by SVM, the classification and regression tree (CART) model, and the *k*-nearest neighbor (k-NN) model. However, Vandal et al. (2019) suggested that the machine learning methods do not provide significant improvements for precipitation downscaling, compared to the conventional statistical methods.

Deep learning methods, which can discover multi-level features in high-dimensional data and capture the potential relationships between multiple environmental variables, are considered as being a significant advance over the conventional machine learning methods (LeCun et al., 2015; Yuan et al., 2020). Among them, the convolutional neural networks (CNNs) can excavate the multi-scale spatial features of high-dimensional images, and are thus widely applied for image super-resolution (Dong et al., 2016; Zhang et al., 2018b; Shen et al., 2020; Qiao et al., 2021; Wu et al., 2022). Numerous studies have demonstrated that CNNs are preeminently adaptable to precipitation downscaling (Baño-Medina et al., 2020; Sun and Lan, 2021; Wang et al., 2021). Consequently, deep learning methods have great potential for remotely sensed precipitation downscaling. However, such applications are still scarce.

Therefore, in this paper, we innovatively propose an attention mechanism based convolutional network (AMCN) to downscale satellite precipitation data. To take full advantage of the complicated characteristics of various materials, an attention mechanism with different modalities is used for the mutual calibration of the feature maps. In addition, multi-level features are extracted by an improved residual convolutional module, to deeply explore the potential relationships between precipitation and ancillary data. Finally, geographic difference analysis (GDA, Duan and Bastiaanssen, 2013; Tan et al., 2022) is introduced to calibrate the remotely sensed observations with in-situ measurements for generating high-quality and fine-scale precipitation images over China.

The rest of this paper is arranged as follows. Section 2 introduces the GPM precipitation dataset and the various ancillary materials. The proposed AMCN method is then detailed in Section 3. The effectiveness of the AMCN method is evaluated and discussed in Section 4. Finally, Section 5 summarizes the contributions, deficiencies, and prospects of this research.

## 2. Datasets

In this study, the monthly satellite-based precipitation data were derived from an Integrated Multi-satellitE Retrievals for the GPM mission (IMERG) dataset with a spatial resolution of 0.1°, which is a data source that has been broadly demonstrated to be more accurate in the publicly released satellite precipitation datasets. In-situ rainfall measurements of 612 meteorological stations were obtained from the China Meteorological Data Service Centre and were used for the accuracy assessment, as shown in Fig. 1.

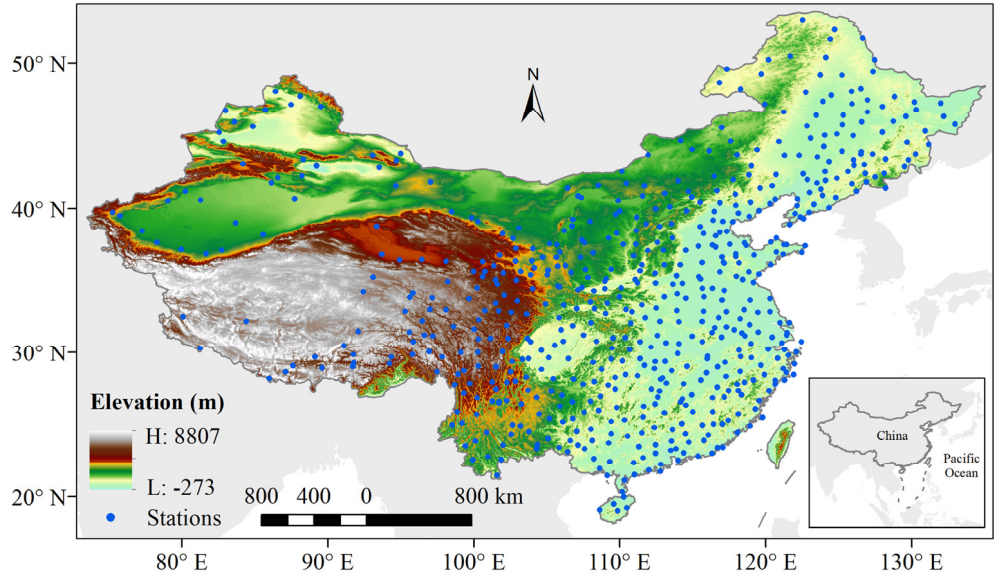

**Fig. 1.** Topographic relief and meteorological stations of China.

For the ancillary remotely sensed datasets, daytime and nighttime land surface temperature (LSTD and LSTN), the enhanced vegetation index (EVI), and land surface reflectance data were derived from the MOD11A2, MOD13A3, and MOD09A1 products of the Moderate Resolution Imaging Spectroradiometer (MODIS), respectively. Digital elevation model (DEM) data were obtained from the Shuttle Radar Topography Mission (STRM). All the ancillary data were resampled to two spatial scales of 0.01° and 0.1° in the WGS84 coordinate system with bilinear interpolation and averaged to a monthly temporal scale. Low-pass filtering (Xu et al., 2015) and inverse distance weighting methods were also applied to identify outliers and fill gaps in the optical data, respectively.

## 3. Methodology

### 3.1. Precipitation downscaling framework

In this study, nine geographic and environmental factors were used to assist the precipitation downscaling. The geographic factors were longitude, latitude, and DEM. The environmental factors were LSTD, LSTN, EVI, the temperature-vegetation dryness index (TVDI), the normalized difference water index (NDWI), and the land surface water index (LSWI), representing surface temperature, vegetation cover, and inland water conditions. The correlation between precipitation and the various ancillary factors can be expressed as:

$$X_P = f(Y_P, Longitude, Latitude, DEM, LSTD, LSTN, EVI, TVDI, NDWI, LSWI) \qquad (1)$$

where $X_P$ and $Y_P$ represent the HR and low-resolution (LR) precipitation data, respectively. The TVDI is an empirical variable widely used in soil moisture inversion, parameterizing the relationship between surface temperature and vegetation index (Sandholt et al., 2002; Gao et al., 2011). It can be formulated as:

$$TVDI = \frac{Ts - Ts_{\min}}{Ts_{\max} - Ts_{\min}} \tag{2}$$

where $Ts$ denotes the land surface temperature (LST) in any target pixel. $Ts_{\min}$ and $Ts_{\max}$ represent the minimum LST on the wet edge and maximum LST on the dry edge of the "triangular" space, respectively. More details can be found in (Sandholt et al., 2002). In addition, the NDWI (Gao, 1996) and LSWI (Chandrasekar et al., 2010) are given as follows:

$$NDWI = \frac{\rho_{nir} - \rho_{mir}}{\rho_{nir} + \rho_{mir}} \tag{3}$$

$$LSWI = \frac{\rho_{nir} - \rho_{swir2}}{\rho_{nir} + \rho_{swir2}} \tag{4}$$

where $\rho_{nir}$, $\rho_{mir}$, and $\rho_{swir2}$ represent the near-infrared, mid-infrared, and short-wave infrared bands of MODIS surface reflectance, respectively.

### 3.2. Attention mechanism based convolutional network

In order to refine the complex characteristics of precipitation with a highly skewed structure, the end-to-end AMCN method is proposed, as shown in Fig. 2. In the network, the LR precipitation ($Y_P$) data and the various fine ancillary data ($X_A$) are used as the input. To take full advantage of the more accurate magnitude information of the coarse precipitation data and the richer details of the fine ancillary data, two cross-attention modules are proposed to extract and recalibrate the feature maps, i.e., a global cross-attention (GCA) module and a multi-factor cross-attention (MFCA) module. Subsequently, a residual densely connected module embedded with an attention mechanism (RDAM) is used to obtain higher-level information from the input feature maps. Finally, the HR precipitation data ($X_P$) are projected through global residual learning, as shown in equation (5):

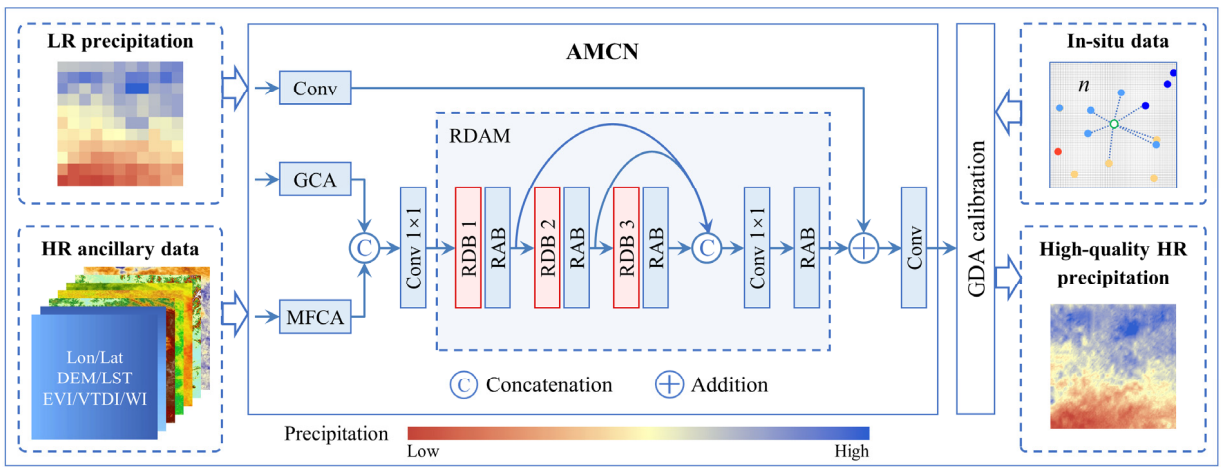

**Fig. 2.** The structure of the AMCN method and the specific schematics for HR precipitation estimation.

$$X_P = f_u(Y_P) + \xi(Y_P, X_A) \qquad (5)$$

where $f_u(\cdot)$ represents the up-sampling function, which is bilinear interpolation in this study. $\xi(\cdot)$ represents the output of the residual network. Furthermore, the loss function made up of Charbonnier loss and degradation loss is proposed to constrain the training procedure. The details of each module in the network are provided below.

### 3.2.1 GCA and MFCA modules

Taking into account the complementary information in the input data, a cross-attention mechanism (Lin et al., 2022a; Lin et al., 2022b) is used to mutually recalibrate the multi-scale feature maps through two different data combination modes, corresponding to the GCA module and the MFCA module.

As illustrated in Fig. 3(a), the cross-attention mechanism consists of a channel attention block for the LR precipitation data (LPCA) and a spatial attention block for the HR ancillary data (HASA). In the LPCA block, the actual magnitude information of the LR precipitation data is used to enhance the feature maps of the HR ancillary data. Specifically, the feature maps containing the magnitude information are extracted from the up-sampled precipitation data through a convolutional layer and then normalized by a sigmoid function to obtain the channel weights. The feature maps extracted from the HR ancillary data can thus be recalibrated by element-wise weighting operations. Due to the highly skewed distribution and wide dynamic range of precipitation, the pooling operation extensively applied in channel attention blocks is removed to avoid data anomalies. Correspondingly, the spatial weights derived from the HR ancillary data are used to enhance the detailed characteristics of the LR precipitation data. Finally, the channel-weighted and spatially weighted feature maps generated by the LPCA and HASA blocks are fused by an element-wise addition operation, to obtain mutually recalibrated feature maps containing abundant information of magnitude and variability. The cross-attention mechanism can be formulated as follows:

$$F_{HASA} = (W_{HASA1} \circ F_P + b_{HASA1}) \otimes \sigma(W_{HASA2} \circ F_A + b_{HASA2}) \qquad (6)$$

$$F_{LPCA} = (W_{LPCA1} \circ F_A + b_{LPCA1}) \otimes \sigma(W_{LPCA2} \circ F_P + b_{LPCA2}) \qquad (7)$$

$$F_{CroA} = F_{HASA} + F_{LPCA} \qquad (8)$$

where $F_{HASA}$ and $F_{LPCA}$ are the output feature maps of the HASA and LPCA blocks, respectively. $F_A$ and $F_P$ denote the input feature maps extracted from the input ancillary data and precipitation data, respectively. $W$ denotes the convolution kernel in the cross-attention block. $\otimes$ and $\circ$ represent the element-wise multiplication operation and the convolution operation, respectively. $b$ is the bias of each attention block. The output feature maps of the cross-attention block ($F_{CroA}$) are then generated by element-wise addition of $F_{HASA}$ and $F_{LPCA}$.

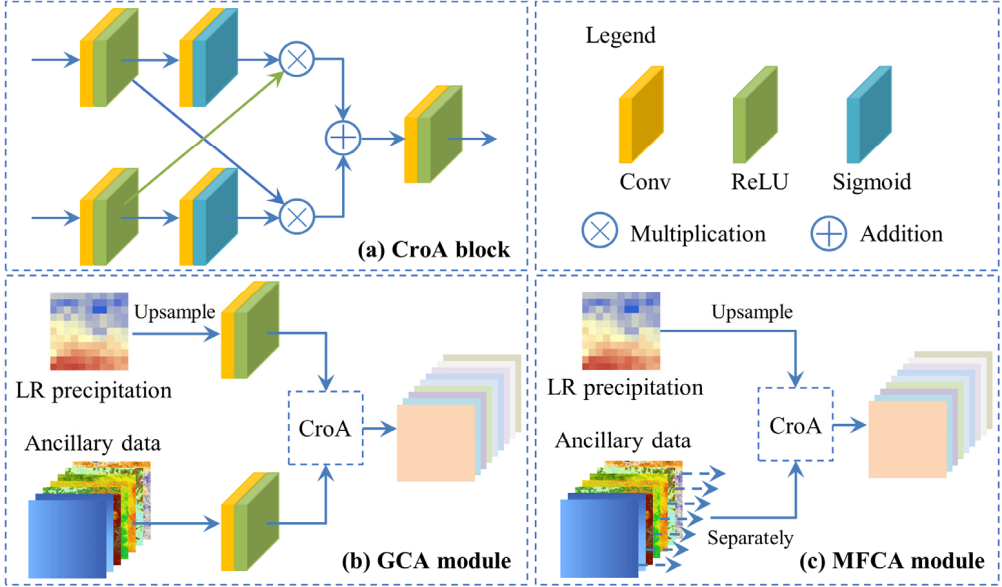

**Fig. 3.** The two cross-attention modules for feature extraction. (a) The cross-attention (CroA) block. (b) The GCA module. (c) The MFCA module.

In terms of the GCA module, as shown in Fig. 3(b), the feature maps extracted from the LR precipitation data and integrated ancillary data are mutually recalibrated in a cross-attention block. Due to the distinct characteristics of the various ancillary data, the GCA module is inevitably incomprehensive. Therefore, the MFCA module is designed to further enrich the feature maps. In this module, the spatial information of each ancillary data is separately fused with the channel information of the precipitation data to obtain a series of recalibrated feature maps. As a result, this module contains nine single-factor cross-attention (SFCA) blocks. Subsequently, the feature maps extracted from the two cross-attention modules are combined through a cascade function as the input of the subsequent RDAM module. The improvement in the proposed network brought by the two cross-attention modules is discussed in Section 4.2.1.

### 3.2.2. RDAM module

The RDAM module is designed to further extract multi-level information from the input feature maps. As illustrated in Fig. 2, the RDAM module consists of a three-level structure, with each level containing a residual dense block (RDB, Zhang et al., 2018b) and a residual attention block (RAB). The three-level feature maps are globally fused by concatenation to improve the retention of the multi-level information flow. In addition, the local features are fully utilized through six densely connected layers in each RDB. However, since the input feature maps are extracted from a variety of input data with complex characteristics, and simple RDBs cannot adequately capture feature heterogeneity in high-dimensional feature maps, the RABs are embedded to continually enhance the feature maps in the proposed approach. As shown in Fig. 4, the channel attention and spatial attention mechanisms are used to recalibrate all the feature maps extracted by the previous modules. Specifically, the channel attention sub-block includes an average pooling operation, a fully connected layer, and a sigmoid normalization operation. The spatial attention sub-block consists of a convolutional layer and a sigmoid normalization operation. Subsequently, the two sub-blocks are combined by an element-wise addition operation. Finally, the output feature maps of each RAB are generated through local residual learning, as shown in equation (9):

$$F_{RAB} = F_{input} + (W_{RAB} \circ (F_{CA} + F_{SA}) + b_{RAB}) \tag{9}$$

where $F_{input}$ and $F_{RAB}$ are the input and output feature maps of the RAB, respectively. $F_{CA}$ and $F_{SA}$ represent the output feature maps of the channel attention and spatial attention sub-blocks, respectively. $W_{RAB}$ and $b_{RAB}$ represent the convolution kernels and bias terms of the RAB, respectively. The RDAM module can extract and recalibrate the multi-level magnitude and variability information through the RDBs and RABs. Finally, the HR precipitation data are estimated by global residual learning.

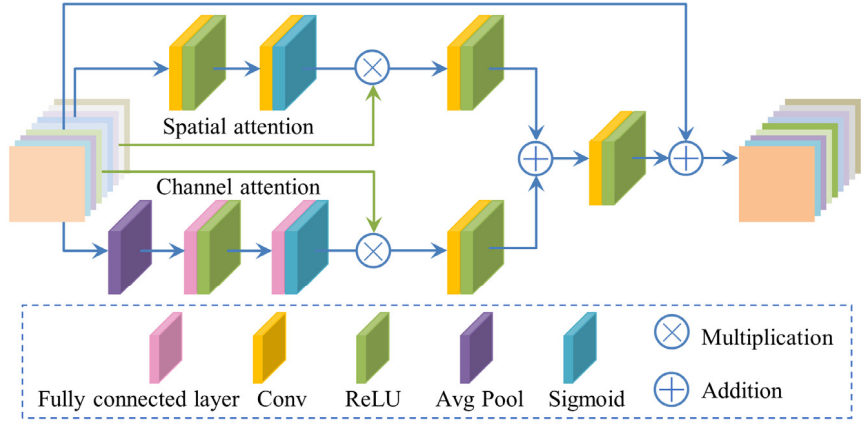

**Fig. 4.** The residual attention block (RAB).

### 3.2.3. Loss functions

The predicted HR precipitation data should be physically consistent with the input LR precipitation data. Therefore, based on the image scale transformation model, a degradation loss function $L_d(\Theta)$ is proposed to physically constrain the training process. This can be formulated as:

$$L_d(\Theta) = \frac{1}{N} \sum_{i=1}^{N} \sqrt{\left\| y_P^i - f_d(f_u(y_P^i) + \xi(y_P^i, x_A^i)) \right\|^2 + \varepsilon^2} \tag{10}$$

where $y_P^i$ and $x_A^i$ represent the input LR precipitation and HR ancillary data, respectively. $f_d(\cdot)$ represents the down-sampling function with bilinear interpolation. As mentioned above, $\xi(\cdot)$ represents the output of the residual network. $\varepsilon$ is the constant term, which is empirically set to 0.001. $N$ is the number of training patches. The degradation loss function constrains the training process at the coarse scale to obtain a more robust performance under different time and scale variations. The effectiveness of the degradation loss function is further discussed in Section 4.2.2.

In addition, a Charbonnier loss function (Lai et al., 2017) is introduced to constrain the network training at the fine scale. Since precipitation data are characterized by a highly skewed distribution and wide dynamic range, the Charbonnier loss $L_c(\Theta)$ is superior to the $L_2$ loss, due to its insensitivity to outliers. The formula for the Charbonnier loss function can be expressed as:

$$L_c(\Theta) = \frac{1}{N} \sum_{i=1}^{N} \sqrt{\left\| \rho^i - \xi(y_P^i, x_A^i) \right\|^2 + \varepsilon^2} \tag{11}$$

where $\rho^i$ represents the residuals between the actual HR precipitation data $\hat{x}_P^i$ and the up-sampled LR precipitation data $f_u(y_P^i)$, $\rho^i = \hat{x}_P^i - f_u(y_P^i)$. Consequently, the total loss function is defined as:

$$L_{total}(\Theta) = \alpha L_c(\Theta) + \beta L_d(\Theta) \tag{12}$$

$$\alpha = \frac{L_c(\Theta)}{L_c(\Theta) + L_d(\Theta)}, \beta = \frac{L_d(\Theta)}{L_c(\Theta) + L_d(\Theta)} \tag{13}$$

where $\alpha$ and $\beta$ are the regularization parameters for the tradeoff between items, which are adaptively determined by $L_c(\Theta)$ and $L_d(\Theta)$. The trained network with a converged total loss is then used for the downscaling testing.

## 4. Results and discussion

### 4.1. Evaluation of precipitation downscaling

To demonstrate the effectiveness of the proposed AMCN method for HR precipitation estimation, simulated and real-data downscaling experiments were conducted, and are described in this section. The proposed method was compared with RF (Breiman, 2001), a back-propagation neural network (BPNN, Rumelhart et al., 1986), and the MARS model (Tan et al., 2022) in both experiments. Three quantitative metrics are employed here for the accuracy assessment with respect to the magnitude and variability of the precipitation data, i.e., the determination coefficient ($R^2$), bias (i.e., result − truth), and root-mean-square error (RMSE). The specific experiments were designed as follows.

1) Training data: As listed in Table 1, 12 groups of data from 2018 were used for the network training, each containing precipitation data at a 1° resolution (down-sampled from the original precipitation data) and ancillary data at a 0.1° resolution as the input dataset, with precipitation data at a 0.1° resolution as the label dataset. A bilinear interpolation algorithm was used for the image resampling. Specifically, all out-of-study and optically cloud-covered regions were excluded to improve the pertinence and precision of the network training. The patch sizes of the inputs and labels were 40×40×10 and 40×40×1, respectively. A total of 28288 patches were generated for the training process.
2) Test data: 24 groups of images from 2019 and 2020 were used for the network testing in both the simulated and real-data experiments. The image resolution in the real-data experiment was 10 times that in the training process and simulated experiment.
3) Parameter settings: The Adam optimizer (Kingma and Ba, 2015) was applied as the gradient

descent optimization algorithm. There were 100 epochs in the training process, each containing 441 iterations. The learning rate was initialized as 0.001 and then halved every 50 epochs. The AMCN network was trained under the PyTorch framework in a Windows 10 environment, with an NVIDIA GeForce RTX 2080 SUPER GPU.

**Table 1.** Details of the training and test data. Note: P and A represent the precipitation and ancillary data, respectively. * The number of input data types.

| Experiment | Data size | Temporal span | Resolution Input (P, A) | Resolution Output (P) |
|---|---|---|---|---|
| Training | 12×360×620(×10)* | Jan 2018 – Dec 2018 | 1°, 0.1° | 0.1° |
| Simulated test | 24×360×620(×10) | Jan 2019 – Dec 2020 | 1°, 0.1° | 0.1° |
| Real-data test | 24×3600×6200(×10) | Jan 2019 – Dec 2020 | 0.1°, 0.01° | 0.01° |

**4.1.1. Simulated experiment**

In the simulated experiment, the original satellite precipitation images were upscaled to simulate the LR precipitation data. The simulated precipitation at a 1° resolution and the various ancillary data at a 0.1° resolution were then utilized to estimate finer precipitation at a 0.1° resolution through the trained models. Consequently, the original satellite precipitation images could be considered as references to evaluate the downscaled results with respect to both the visual effects and quantitative metrics, which is called image-based evaluation. In addition, quantitative evaluation with in-situ measurements was also conducted, which is called station-based evaluation.

The downscaled results of the RF, BPNN, MARS, and AMCN methods for February 2019 and July 2020 and their spatial details are presented in Fig. 5 and Fig. 6, respectively. Since the southeast monsoon brings water vapor from the Pacific Ocean, the southeast region has the most abundant precipitation in China. As shown in Fig. 5, this phenomenon is reflected in the two original precipitation images for February 2019 (a1) and July 2020 (b1). The BPNN-, MARS-, and AMCN-downscaled results have spatial patterns that are consistent with the corresponding original images, while the RF-downscaled results present considerable differences for February 2019. Specifically, the input precipitation data (a2) contain very little information, and thus the RF algorithm without sufficient spatial constraints inappropriately introduces the spatial distribution from the various ancillary data. In contrast, the BPNN, MARS, and AMCN methods can effectively extract the spatial characteristics from the coarsest images and retrieve fine-scale precipitation data with a plausible distribution. For July 2020, the downscaled results of all the methods are relatively consistent with the real satellite precipitation data.

The enlarged views in Fig. 6 further reveal the downscaling differences between the different methods. The RF method narrows the spatial dynamic range of the precipitation, resulting in overestimation of the low values and underestimation of the high values. The BPNN- and MARS-downscaled results are relatively smooth, thus attenuating the spatial heterogeneity of the precipitation. Due to the low correlation between the geographic/environmental ancillary factors and precipitation, the BPNN and MARS methods have difficulty in fitting the complicated nonlinear relationships. Meanwhile, the AMCN-downscaled results present abundant spatial details. This reveals that the proposed AMCN method can effectively introduce detailed information from the ancillary data sources, owing to the effectiveness of the attention mechanisms and the feature richness of the deep convolutional filters. In addition, the AMCN method optimizes the anomalous details in the original precipitation data, such as the "orthogon" apparent in Fig. 6(a1). Consequently, the AMCN method achieves more robust visual effects than the other three methods. It should be noted that the RF method was not optimized in this study. In other cases, with sufficient spatial constraints and stable precipitation patterns, the RF can perform well.

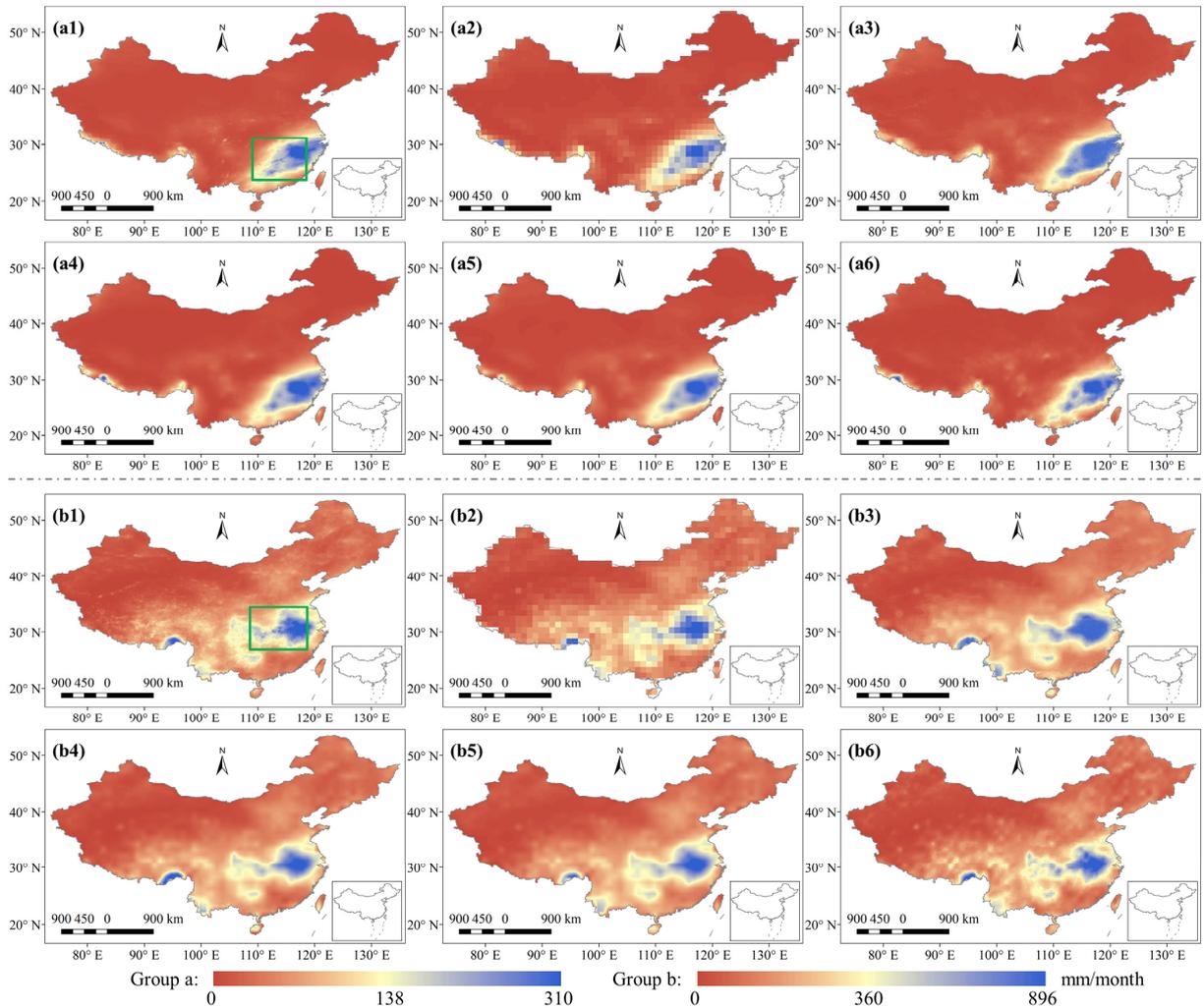

**Fig. 5.** The downscaled precipitation for February 2019 (group a) and July 2020 (group b). (a1)–(a6) and (b1)–(b6) are the original images (references), the input precipitation data, the RF-downscaled results, the BPNN-downscaled results, the MARS-downscaled results, and the AMCN-downscaled results, respectively.

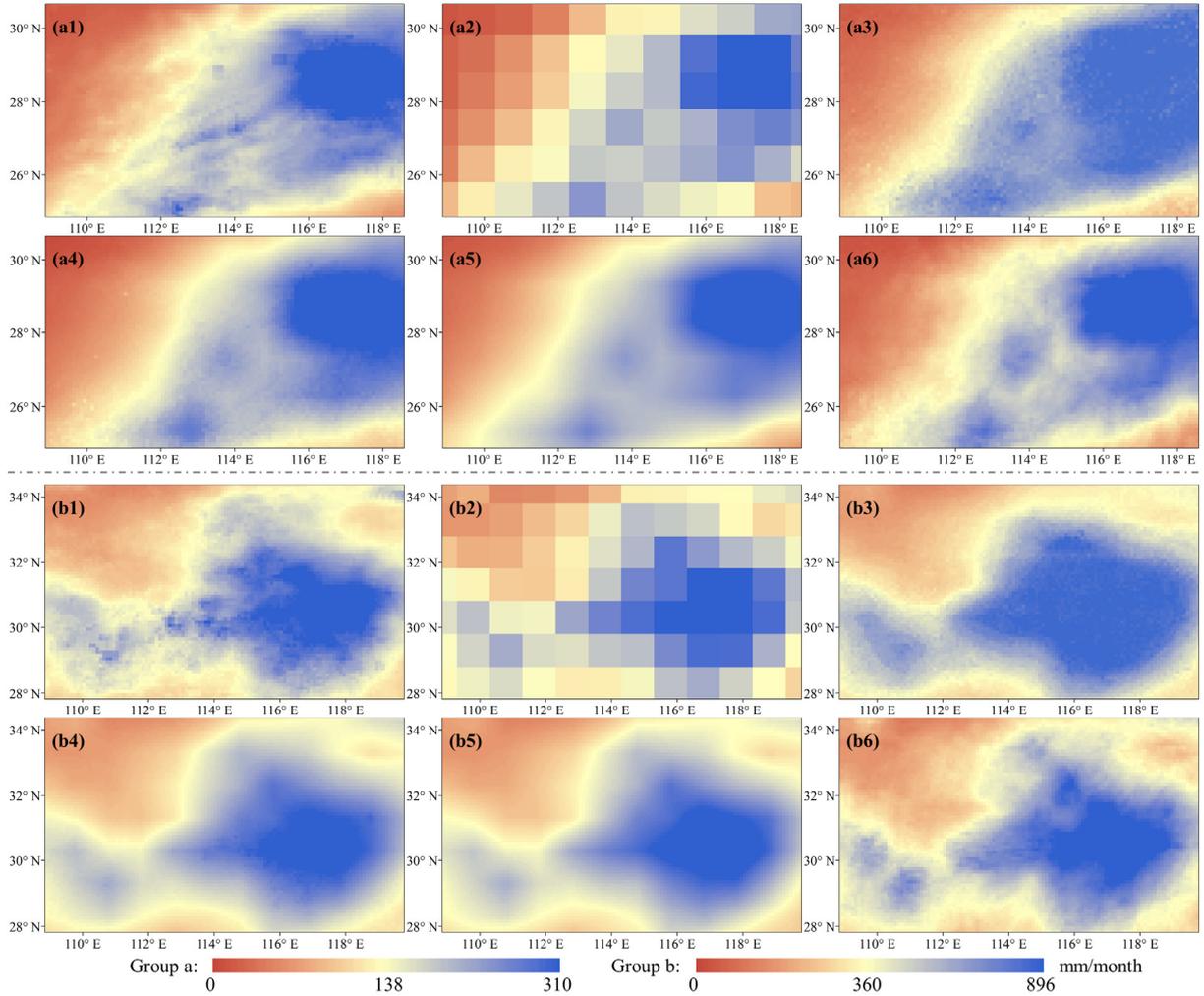

**Fig. 6.** Spatial details of the downscaled precipitation for February 2019 (group a) and July 2020 (group b), corresponding to Fig. 5.

For an in-depth assessment, the downscaled results were quantitatively evaluated based on the original precipitation images and in-situ measurements, as shown in Table 3. In terms of the numerical correlation, the RF-, BPNN-, MARS-, and AMCN-downscaled results all have relatively high consistency with the original images, with $R^2$ values of 0.92, 0.93, 0.92, and 0.95, respectively. Despite the differences in the visual effects, the quantitative metrics for the three baseline methods are similar. However, the AMCN-downscaled results show significantly lower bias and RMSE, indicating that the spatial details are relatively accurate. For the station-based evaluation, the accuracy of the original precipitation images is first given to measure the difference between the satellite observations and the in-situ measurements. It can be seen that the original images have an $R^2$ value of 0.83 and an RMSE of 37.45 mm/month. Among the four methods, the quantitative metrics for the AMCN method are more consistent with the original images, with an $R^2$ and RMSE of 0.80 and 41.67 mm/month, respectively. The high positive biases reveal that the satellite-based precipitation data are overestimated, compared with the in-situ measurements. In addition, Fig. 7 shows the spatial distribution of the quantitative metrics for the AMCN-downscaled time series. The $R^2$ values are prominently high in most areas, compared to the original precipitation images (Fig. 7a). The precipitation in western China is extremely low, and the false detection of light rainfall can clearly reduce the $R^2$ value, but does not significantly affect the RMSE. In contrast, the precipitation in southeastern China is extremely high, and the deviation of the extreme precipitation projection can significantly increase the RMSE. In addition, the RMSE in fringe areas is relatively high, due to the fewer spatial constraints. The spatial distribution of the station-based quantitative metrics is generally consistent with that of the image-based evaluations. In particular, the $R^2$ is significantly lower in fringe areas, since the original images are markedly different from the in-situ measurements. Overall, the quantitative evaluation results further demonstrate the superiority of the AMCN method.

Four methods were used to downscale the precipitation data from 1° to 0.1° in the simulated experiment. The spatial magnitude information retrieved by the RF method is relatively inaccurate. The BPNN and MARS methods have difficulty in effectively retrieving the fine details. Meanwhile, the AMCN-downscaled results present abundant spatial details, which are highly consistent with the original images and in-situ measurements.

**Table 3.** Quantitative evaluation results for the simulated experiment. Unit: mm/month.

| Method | Image-based evaluation | | | Station-based evaluation | | |
|---|---|---|---|---|---|---|
| | $R^2$ | Bias | RMSE | $R^2$ | Bias | RMSE |
| Original | -- | -- | -- | 0.83 | 17.04 | 37.45 |
| RF | 0.92 | 3.40 | 18.28 | 0.75 | 19.53 | 43.67 |
| BPNN | 0.93 | 2.20 | 17.67 | 0.76 | 18.40 | 44.43 |
| MARS | 0.92 | 1.36 | 18.61 | 0.75 | **16.92** | 44.58 |
| AMCN | **0.95** | **0.34** | **14.01** | **0.80** | 18.21 | **41.67** |

Note: The implementation of RF and BPNN was reproduced by ourselves.

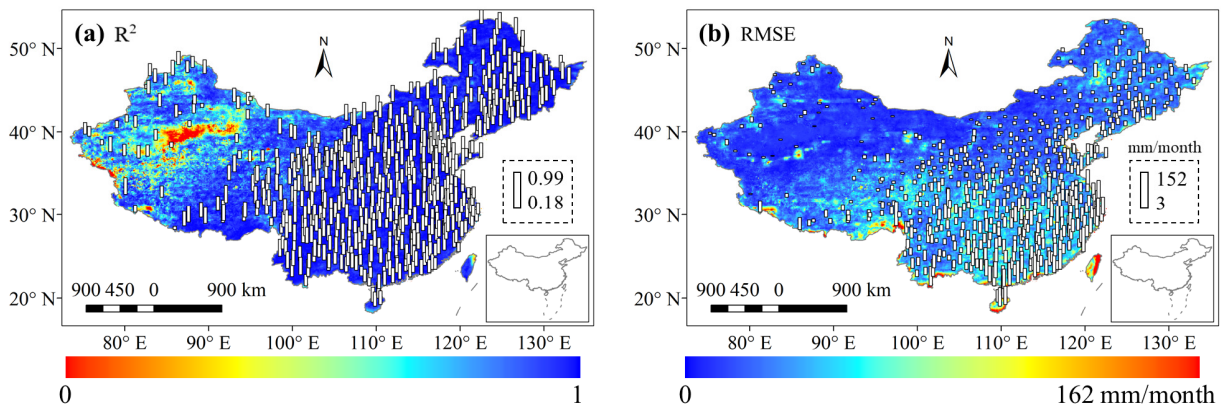

**Fig. 7.** The spatial distribution of the $R^2$ and RMSE for the AMCN-downscaled precipitation data, compared with the original satellite observations and in-situ measurements (columns).

### 4.1.2. Real-data experiment

The original satellite precipitation data were downscaled from 0.1° to 0.01° in the real-data experiment. Subsequently, GDA, which is a residual correction method based on inverse distance weighted interpolation (Duan and Bastiaanssen, 2013; Tan et al., 2022), was introduced to further calibrate the downscaled results with the in-situ measurements. In particular, all the downscaled results with a 0.01° resolution were down-sampled in the image-based quantitative evaluation to match the original precipitation data.

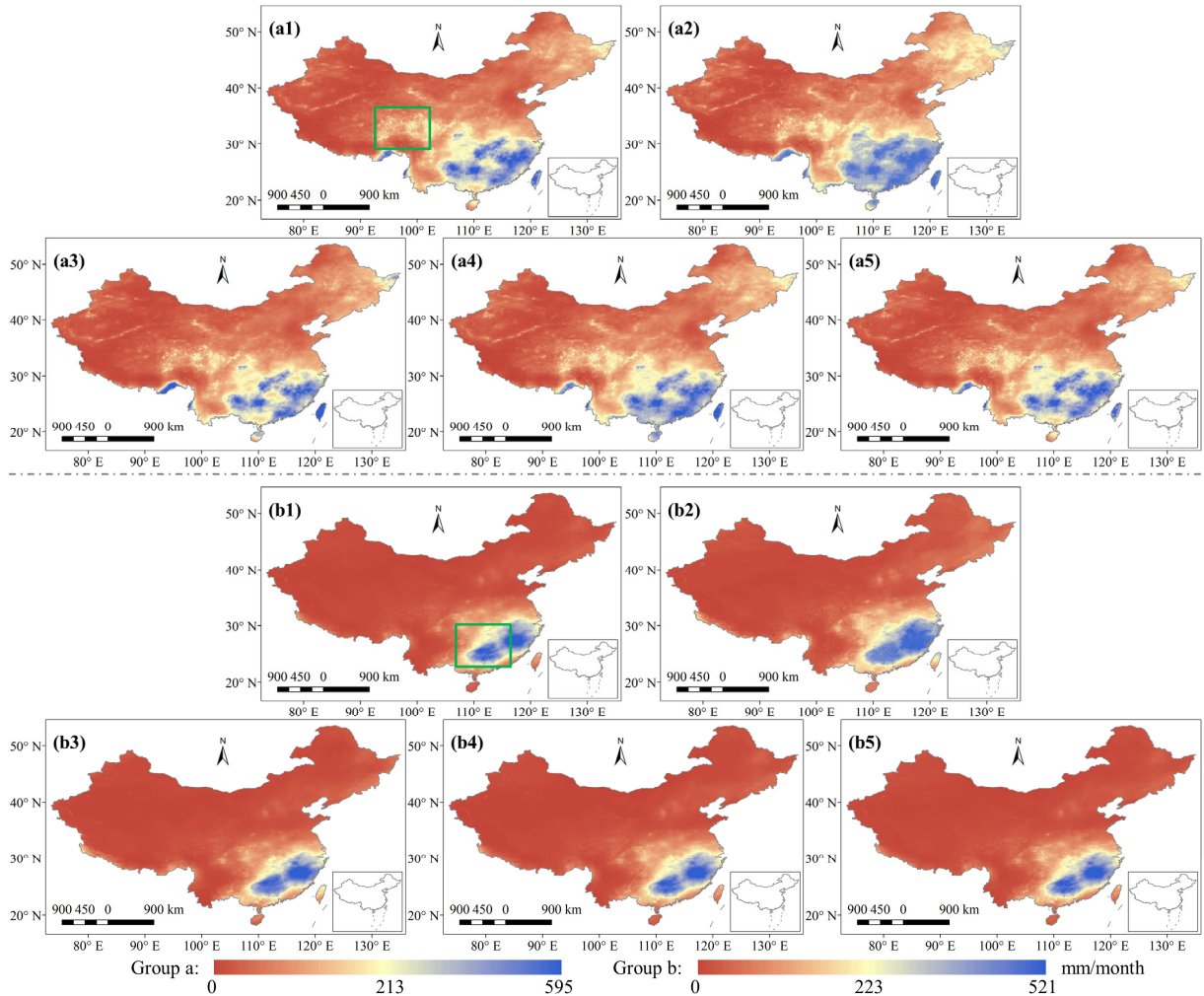

**Fig. 8.** The downscaled precipitation for June 2019 (group a) and March 2020 (group b). (a1)–(a5) and (b1)–(b5) are the original images (references), the RF-downscaled images, the BPNN-downscaled images, the MARS-downscaled images, and the AMCN-downscaled images, respectively.

The downscaling performance of the proposed method was compared with that of the three baseline methods for June 2019 and March 2020, as presented in Fig. 8. The RF-downscaled results still struggle to accurately capture the spatial distribution of the precipitation in the real-data experiment. The BPNN-, MARS-, and AMCN-downscaled results reproduce the spatial patterns well. Only the MARS-downscaled result for the southern region in June 2019 is significantly overestimated.

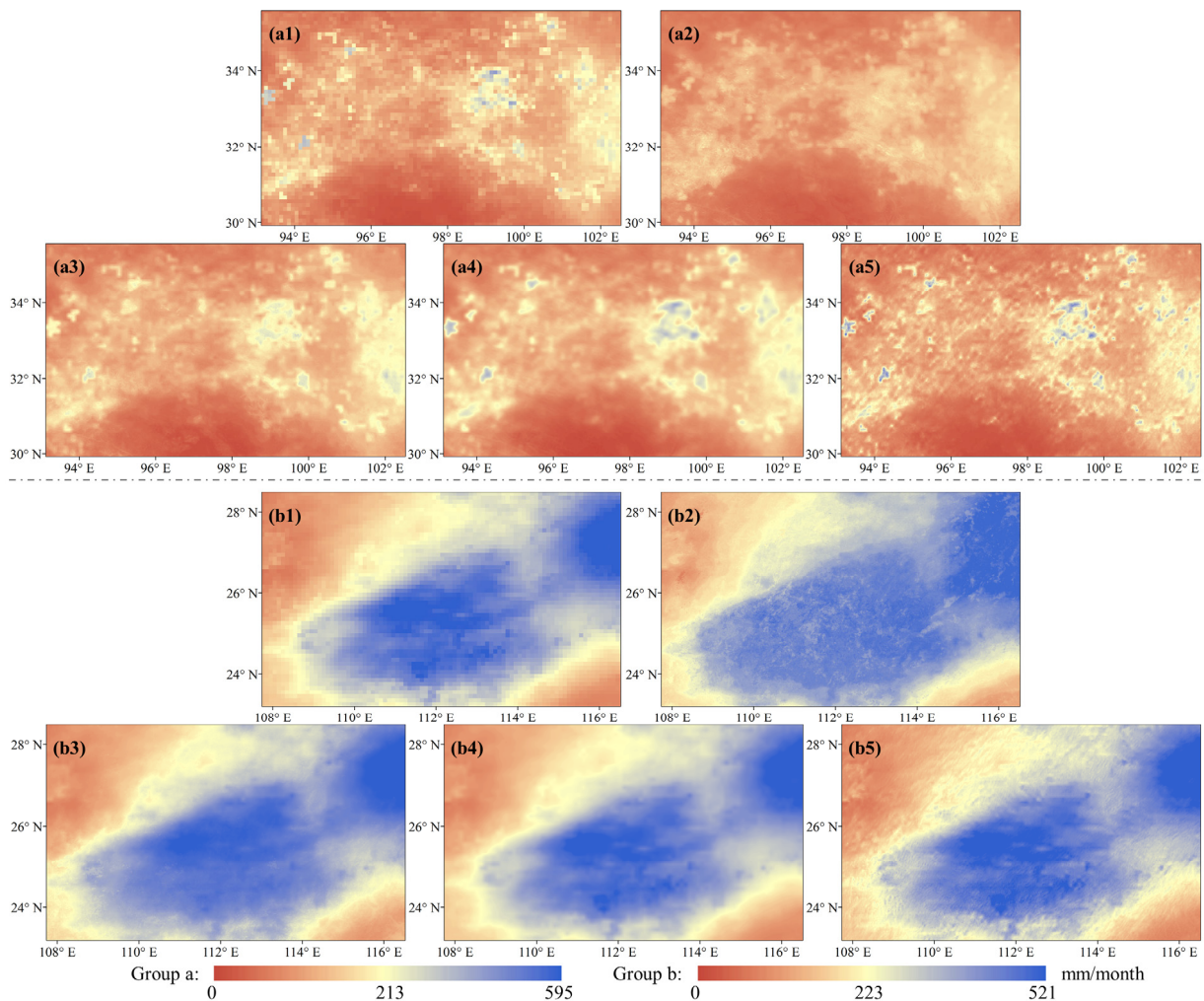

**Fig. 9.** Spatial details of the downscaled precipitation for June 2019 (group a) and March 2020 (group b), corresponding to Fig. 8.

Two different precipitation patterns, i.e., light rainfall and heavy rainfall, are detailed in Fig. 9. The RF-downscaled results generally capture the areal extent but significantly underestimate the intensity for June 2019. The BPNN-, MARS-, and AMCN-downscaled results preserve the spatial distribution in the original images and introduce spatial details from the ancillary data. The difference is that the BPNN- and MARS-downscaled results are smoother, while the AMCN-downscaled results are more detailed. Moreover, the RF method retrieves a larger spatial extent for the high-intensity precipitation and introduces some artifacts for March 2020. The BPNN method also slightly alters the original distribution of the satellite precipitation. The MARS and AMCN methods preserve the original spatial distribution, but the AMCN method retrieves richer details. Therefore, we can conclude that the proposed AMCN method is effective in downscaling original satellite precipitation data.

The quantitative evaluation can reveal the accuracy of the spatial details in the downscaled results. As shown in Table 4, the RF-, BPNN-, MARS-, and AMCN-downscaled results are highly correlated with the original images, with $R^2$ values of greater than 0.94. In terms of the bias and RMSE, the AMCN method is significantly superior to the three baseline methods, demonstrating that its spatial distribution is more consistent with the original images. Subsequently, the downscaled results were assessed based on in-situ measurements. The AMCN method achieves the best performance, followed by the MARS, BPNN, and RF methods. This reveals that the deep convolutional networks generally perform well for precipitation downscaling, whereas the machine learning methods have some uncertainties. The proposed method further decreases the RMSE by 5.41 mm/month, compared to the MARS method, demonstrating the superiority of deep learning methods over statistical regression methods in satellite precipitation downscaling. In addition, as shown in Fig. 10, the quantitative metrics for the AMCN-downscaled results are more evenly distributed in the real-data experiment than in the simulated experiment. The extremes are decreased and are still mainly distributed in fringe areas. This indicates that the proposed AMCN method can robustly estimate HR precipitation data.

**Table 4.** Quantitative evaluation results for the real-data experiment. Unit: mm/month.

| Method | Image-based evaluation | | | Station-based evaluation | | |
|--------|------|------|------|------|------|------|
|        | $R^2$ | Bias | RMSE | $R^2$ | Bias | RMSE |
| RF     | 0.948 | 16.82 | 24.96 | 0.77 | 40.05 | 57.93 |
| BPNN   | 0.962 | 7.02  | 18.71 | 0.79 | 26.95 | 49.84 |
| MARS   | 0.988 | 4.04  | 10.10 | 0.81 | 21.65 | 43.52 |
| AMCN   | **0.996** | **0.27** | **4.23** | **0.82** | **16.99** | **38.11** |

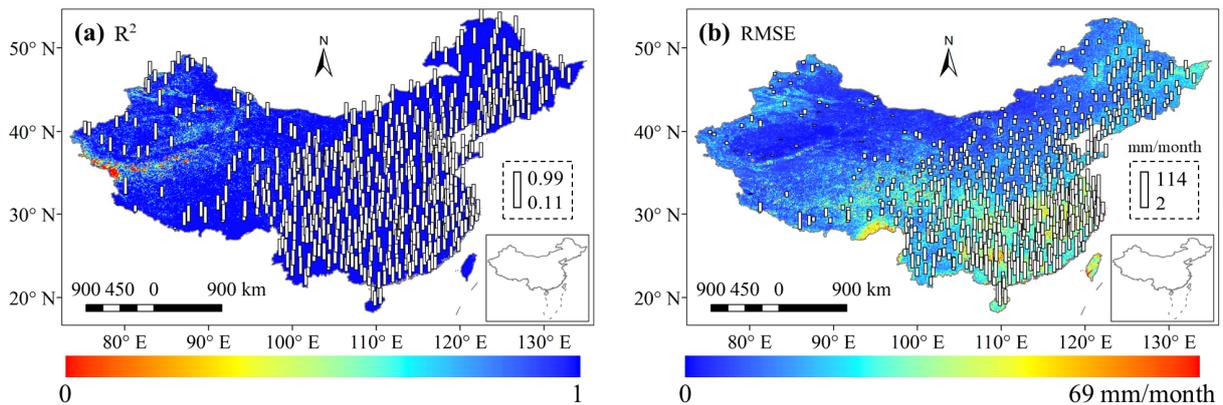

**Fig. 10.** The spatial distribution of the $R^2$ and RMSE for the AMCN-downscaled precipitation data, compared with the original satellite observations and the in-situ measurements (columns).

Since satellite-based precipitation data suffer from inherent biases, compared to in-situ measurements, GDA was introduced to further improve the downscaled data. The calibrated results for June 2020 are shown in Fig. 11. Compared with the in-situ measurements, the original satellite precipitation data are significantly overestimated (Fig. 11a). The GDA method effectively eliminates the systematic biases and retains the fine details of the downscaled results. Nevertheless, both the RF and BPNN methods introduce some artifacts, which cannot be removed by GDA. The MARS-downscaled results still present significant smoothness after calibration. In contrast, the AMCN-GDA procedure can retrieve precipitation data with both accurate magnitude information and fine details. In terms of the quantitative evaluation, all the residual biases are negligible, as shown in Table 5. Although the RMSEs of the three baseline methods are greatly decreased after calibration, the proposed network still exhibits the best performance. This confirms the superiority of the proposed AMCN method for satellite precipitation downscaling, and the potential of the AMCN-GDA procedure to generate high-quality and fine-scale precipitation products.

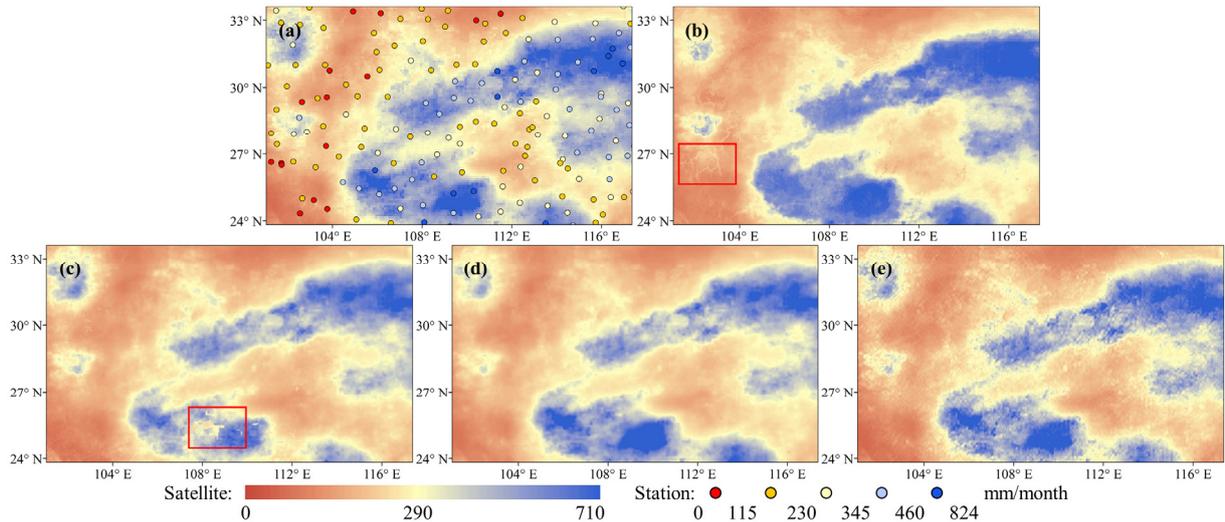

**Fig. 11.** The effectiveness of calibration based on in-situ measurements (colored dots) for June 2020. (a) Original satellite precipitation at a 0.1° resolution and the in-situ measurements. (b)–(e) The calibrated results for the RF-, BPNN-, MARS-, and AMCN-downscaled data.

**Table 5.** Quantitative evaluation for the GDA calibration. Unit: mm/month.

| Method | Station-based evaluation | | |
| --- | --- | --- | --- |
| | $R^2$ | Bias | RMSE |
| RF-GDA | 0.78 | −0.53 | 36.78 |
| BPNN-GDA | 0.79 | −0.09 | 36.92 |
| MARS-GDA | 0.81 | −0.39 | 33.99 |
| AMCN-GDA | **0.83** | −0.46 | **30.88** |

Both the image-based and station-based evaluations indicate that the proposed method obtains significantly higher accuracy than the RF, BPNN, and MARS methods. The AMCN method can obtain downscaled results with an accurate spatial distribution and fine details. Furthermore, the AMCN-GDA procedure is capable of generating high-quality precipitation data at a fine scale.

**4.2. Discussion**

**4.2.1. Ablation experiments with the two cross-attention modules**

To elaborate the effect of the GCA and MFCA modules, ablation experiments were implemented as follows: 1) no cross-attention module; 2) GCA module only; 3) MFCA module only; and 4) both GCA and MFCA modules. In the real-data experiment, the downscaled results for each category are highly consistent with the original images, as shown in Fig. 12. Although the RMSE of Category 1) is not high, its spatial details are insufficient. The performance of the GCA module is poor, due to the insufficient interpretation of the surface characteristics. The MFCA module can adequately extract ancillary features but introduces a few artifacts. Meanwhile, the combination of the GCA and MFCA modules balances the magnitude information and fine details, achieving an optimal RMSE of 4.23 mm/month. Overall, the combination of the two cross-attention modules can improve the robustness of the downscaling process.

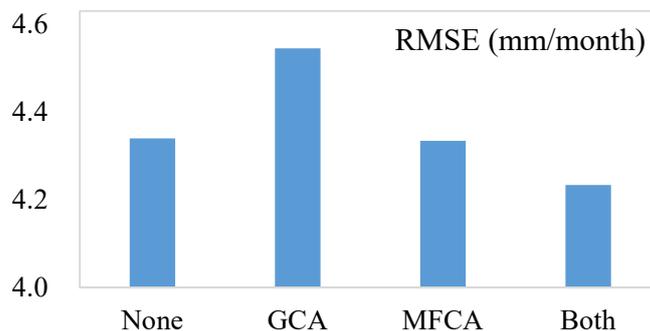

**Fig. 12.** The effectiveness of the two cross-attention modules.

**4.2.2. The effectiveness of the degradation loss function**

In this study, a degradation loss function was designed to physically constrain the network training at a coarse scale. The effectiveness of the degradation loss was further investigated in a supplementary experiment. Due to the stronger spatial constraints from the input precipitation data, the downscaled results of the AMCN method with the degradation loss are more consistent with the original precipitation images. As a result, the image-based evaluation indicates that the degradation loss slightly improves the retention of the magnitude information and reduces the artifacts, with the RMSE decreased from 4.94 to 4.23 mm/month.

**5. Conclusions**

In this study, deep learning technology was innovatively used for satellite precipitation downscaling. In this paper, we have proposed the novel attention mechanism based convolutional network (AMCN) method, which consists of GCA, MFCA, and RDAM modules. The GCA and MFCA modules are two cross-attention mechanisms with different data combinations, which are used to fully exploit the multi-scale complementary information of the LR precipitation data and the various HR ancillary data sources. Subsequently, the RDAM module is proposed to further capture multi-level features for potential relationship exploration. In addition, a degradation loss function based on LR precipitation data is designed to reduce artifacts and improve the robustness of the network. In both simulated and real-data experiments with data from China, the AMCN-downscaled results are more consistent with the actual satellite images and in-situ measurements than the RF-, BPNN-, and MARS-downscaled results. The RF method cannot accurately retrieve the spatial distribution of heavy precipitation. The BPNN and MARS methods cannot effectively reflect the fine spatial details, due to their smoothness. Meanwhile, the proposed AMCN method can estimate precipitation with both an accurate spatial distribution and abundant fine details, thanks to the deep convolutional layers and the two attention modules. The AMCN method was also shown to be superior to the three baseline methods in the quantitative evaluations, revealing that its detailed information is reliable. These results demonstrate the effectiveness of the proposed AMCN method for downscaling satellite-based precipitation data. Finally, the downscaled results were further calibrated with the GDA method for high-quality and fine-scale precipitation estimation over China.

Nevertheless, the proposed method is restricted by the spatio-temporal resolution of the ancillary optical data. Provided that various ancillary data sources with a high spatio-temporal resolution and continuity are available, this method represents a promising approach for downscaling precipitation data with a higher temporal resolution (daily or sub-daily), which is a potential research area.

## References

Baño-Medina, J., Manzanas, R., and Gutierrez, J. M., 2020. Configuration and intercomparison of deep learning neural models for statistical downscaling, Geosci. Model Dev. 13(4), 2109-2124, https://doi.org/10.5194/gmd-13-2109-2020.

Boussetta, S., Koike, T., Yang, K., Graf, T., and Pathmathevan, M., 2008. Development of a coupled land-atmosphere satellite data assimilation system for improved local atmospheric simulations, Remote Sens. Environ. 112(3), 720-734, https://doi.org/10.1016/j.rse.2007.06.002.

Breiman, L., 2001. Random forests, Machine Learning 45(1), 5-32, https://doi.org/10.1023/A:1010933404324.

Chandrasekar, K., Sesha Sai, M. V. R., Roy, P. S., and Dwevedi, R. S., 2010. Land Surface Water Index (LSWI) response to rainfall and NDVI using the MODIS vegetation index product, Int. J. Remote Sens. 31(15), 3987-4005, https://doi.org/10.1080/01431160802575653.

Chaudhuri, C., and Srivastava, R., 2017. A novel approach for statistical downscaling of future precipitation over the Indo-Gangetic Basin, J. Hydrol. 547, 21-38, https://doi.org/10.1016/j.jhydrol.2017.01.024.

Chen, C., Zhao, S., Duan, Z., and Qin, Z., 2015. An Improved Spatial Downscaling Procedure for TRMM 3B43 Precipitation Product Using Geographically Weighted Regression, IEEE J. Sel. Top. Appl. Earth Obs. Remote Sens. 8(9), 4592-4604, https://doi.org/10.1109/JSTARS.2015.2441734.

Chen, C., Hu, B., and Li, Y., 2021. Easy-to-use spatial random-forest-based downscaling-calibration method for producing precipitation data with high resolution and high accuracy, Hydrol. Earth Syst. Sci. 25(11), 5667-5682, https://doi.org/10.5194/hess-25-5667-2021.

Chen, F. R., Gao, Y. Q., Wang, Y. G., and Li, X., 2020. A downscaling-merging method for high-resolution daily precipitation estimation, J. Hydrol. 581, 124414, https://doi.org/10.1016/j.jhydrol.2019.124414.

Dong, C., Loy, C. C., He, K., and Tang, X., 2016. Image Super-Resolution Using Deep Convolutional Networks, IEEE Trans. Pattern Anal. Mach. Intell. 38(2), 295-307, https://doi.org/10.1109/TPAMI.2015.2439281.

Duan, Z., and Bastiaanssen, W. G. M., 2013. First results from Version 7 TRMM 3B43 precipitation product in combination with a new downscaling-calibration procedure, Remote Sens. Environ. 131, 1-13, https://doi.org/10.1016/j.rse.2012.12.002.

Gao, B. C., 1996. NDWI - A normalized difference water index for remote sensing of vegetation liquid water from space, Remote Sens. Environ. 58(3), 257-266, https://doi.org/10.1016/S0034-4257(96)00067-3.

Gao, Z., Gao, W., and Chang, N. B., 2011. Integrating temperature vegetation dryness index (TVDI) and regional water stress index (RWSI) for drought assessment with the aid of LANDSAT TM/ETM+images, Int. J. Appl. Earth Obs. Geoinf. 13(3), 495-503, https://doi.org/10.1016/j.jag.2010.10.005.

Ghorbanpour, A. K., Hessels, T., Moghim, S., and Afshar, A., 2021. Comparison and assessment of spatial downscaling methods for enhancing the accuracy of satellite-based precipitation over Lake Urmia Basin, J. Hydrol. 596, 126055, https://doi.org/10.1016/j.jhydrol.2021.126055.

Immerzeel, W. W., Rutten, M. M., and Droogers, P., 2009. Spatial downscaling of TRMM precipitation using vegetative response on the Iberian Peninsula, Remote Sens. Environ. 113(2), 362-370, https://doi.org/10.1016/j.rse.2008.10.004.

Jia, S., Zhu, W., Lu, A., and Yan, T., 2011. A statistical spatial downscaling algorithm of TRMM precipitation based on NDVI and DEM in the Qaidam Basin of China, Remote Sens. Environ. 115(12), 3069-3079, https://doi.org/10.1016/j.rse.2011.06.009.

Jing, W., Yang, Y., Yue, X., and Zhao, X., 2016. A comparison of different regression algorithms for downscaling monthly satellite-based precipitation over North China, Remote Sensing 8(10), 835, https://doi.org/10.3390/rs8100835.

Kingma, D. P., and Ba, J. L., 2015. Adam: A method for stochastic optimization, 3rd International Conference on Learning Representations, ICLR 2015 - Conference Track Proceedings, 1-11, https://doi.org/10.48550/arXiv.1412.6980.

Lai, W. S., Huang, J. B., Ahuja, N., Yang, M. H., and Ieee, 2017. Deep Laplacian Pyramid Networks for Fast and Accurate Super-Resolution, 30th IEEE Conference on Computer Vision and Pattern Recognition (CVPR 2017), WOS:000418371405097, 5835-5843, https://doi.org/10.1109/CVPR.2017.618.

LeCun, Y., Bengio, Y., and Hinton, G., 2015. Deep learning, Nature 521(7553), 436-444, https://doi.org/10.1038/nature14539.

Lin, L., Li, J., Shen, H., Zhao, L., Yuan, Q., and Li, X., 2022a. Low-Resolution Fully Polarimetric SAR and High-Resolution Single-Polarization SAR Image Fusion Network, IEEE Trans. Geosci. Remote Sens. 60, 1-17, https://doi.org/10.1109/TGRS.2021.3121166.

Lin, L., Shen, H., Li, J., and Yuan, Q., 2022b. FDFNet: A Fusion Network for Generating High-Resolution Fully PolSAR Images, IEEE Geosci. Remote Sens. Lett. 19, 1-5, https://doi.org/10.1109/LGRS.2021.3127958.

Lincoln, T., 2008. Climate science: A bright side of precipitation, Nature 455(7211), 298-298, https://doi.org/10.1038/455298a.

Ma, Z. G., Xu, J. T., He, K., Han, X. Z., Ji, Q. W., Wang, T. C., Xiong, W. T., and Hong, Y., 2020. An updated moving window algorithm for hourly-scale satellite precipitation downscaling: A case study in the Southeast Coast of China, J. Hydrol. 581, 124378, https://doi.org/10.1016/j.jhydrol.2019.124378.

Madakumbura, G. D., Thackeray, C. W., Norris, J., Goldenson, N., and Hall, A., 2021. Anthropogenic influence on extreme precipitation over global land areas seen in multiple observational datasets, Nat. Commun. 12(1), 1-9, https://doi.org/10.1038/s41467-021-24262-x.

Qiao, C., Li, D., Guo, Y., Liu, C., Jiang, T., Dai, Q., and Li, D., 2021. Evaluation and development of deep neural networks for image super-resolution in optical microscopy, Nat. Methods 18(2), 194-202, https://doi.org/10.1038/s41592-020-01048-5.

Rumelhart, D. E., Hinton, G. E., and Williams, R. J., 1986. Learning representations by back-propagating errors, nature 323(6088), 533-536, https://doi.org/10.1038/323533a0.

Sandholt, I., Rasmussen, K., and Andersen, J., 2002. A simple interpretation of the surface temperature/vegetation index space for assessment of surface moisture status, Remote Sens. Environ. 79(2-3), 213-224, https://doi.org/10.1016/S0034-4257(01)00274-7.

Sarojini, B. B., Stott, P. A., and Black, E., 2016. Detection and attribution of human influence on regional precipitation, Nat. Clim. Change 6(7), 669-675, https://doi.org/10.1038/nclimate2976.

Shen, H., Lin, L., Li, J., Yuan, Q., and Zhao, L., 2020. A residual convolutional neural network for polarimetric SAR image super-resolution, ISPRS J. Photogramm. Remote Sens. 161, 90-108, https://doi.org/10.1016/j.isprsjprs.2020.01.006.

Shen, Z., and Yong, B., 2021. Downscaling the GPM-based satellite precipitation retrievals using gradient boosting decision tree approach over Mainland China, J. Hydrol. 602, 126803, https://doi.org/10.1016/j.jhydrol.2021.126803.

Sorooshian, S., Aghakouchak, A., Arkin, P., Eylander, J., Foufoula-Georgiou, E., Harmon, R., Hendrickx, J. M. H., Imam, B., Kuligowski, R., Skahill, B., and Skofronick-Jackson, G., 2011. Advancing the remote sensing of precipitation, Bull. Am. Meteorol. Soc. 92(10), 1271-1272, https://doi.org/10.1175/BAMS-D-11-00116.1.

Sun, L., and Lan, Y., 2021. Statistical downscaling of daily temperature and precipitation over China using deep learning neural models: Localization and comparison with other methods, Int. J. Climatol. 41(2), 1128-1147, https://doi.org/10.1002/joc.6769.

Tan, W., Tian, L., Shen, H., and Zeng, C., 2022. A New Downscaling-Calibration Procedure for TRMM Precipitation Data Over Yangtze River Economic Belt Region Based on a Multivariate Adaptive Regression Spline Model, IEEE Trans. Geosci. Remote Sens. 60, 1-19, https://doi.org/10.1109/TGRS.2021.3087896.

Vandal, T., Kodra, E., and Ganguly, A. R., 2019. Intercomparison of machine learning methods for statistical downscaling: the case of daily and extreme precipitation, Theor. Appl. Climatol. 137(1-2), 557-570, https://doi.org/10.1007/s00704-018-2613-3.

Wang, F., Tian, D., Lowe, L., Kalin, L., and Lehrter, J., 2021. Deep Learning for Daily Precipitation and Temperature Downscaling, Water Resour. Res. 57(4), e2020WR029308, https://doi.org/10.1029/2020WR029308.

Wang, L., Chen, R., Han, C., Yang, Y., Liu, J., Liu, Z., Wang, X., Liu, G., and Guo, S., 2019. An improved spatial-temporal downscaling method for TRMM precipitation datasets in alpine regions: A case study in northwestern China's Qilian Mountains, Remote Sensing 11(7), 870, https://doi.org/10.3390/RS11070870.

Wu, J., Lin, L., Li, T., Cheng, Q., Zhang, C., and Shen, H., 2022. Fusing Landsat 8 and Sentinel-2 data for 10-m dense time-series imagery using a degradation-term constrained deep network, Int. J. Appl. Earth Obs. Geoinf. 108, 102738, https://doi.org/10.1016/j.jag.2022.102738.

Xu, S., Wu, C., Wang, L., Gonsamo, A., Shen, Y., and Niu, Z., 2015. A new satellite-based monthly precipitation downscaling algorithm with non-stationary relationship between precipitation and land surface characteristics, Remote Sens. Environ. 162, 119-140, https://doi.org/10.1016/j.rse.2015.02.024.

Yan, X., Chen, H., Tian, B., Sheng, S., Wang, J., and Kim, J. S., 2021. A downscaling–merging scheme for improving daily spatial precipitation estimates based on random forest and cokriging, Remote Sensing 13(11), 2040, https://doi.org/10.3390/rs13112040.

Yuan, Q., Shen, H., Li, T., Li, Z., Li, S., Jiang, Y., Xu, H., Tan, W., Yang, Q., Wang, J., Gao, J., and Zhang, L., 2020. Deep learning in environmental remote sensing: Achievements and challenges, Remote Sens. Environ. 241, 111716, https://doi.org/10.1016/j.rse.2020.111716.

Zeng, Z., Chen, H., Shi, Q., and Li, J., 2022. Spatial Downscaling of IMERG Considering Vegetation Index Based on Adaptive Lag Phase, IEEE Trans. Geosci. Remote Sens. 60, 1-15, https://doi.org/10.1109/TGRS.2021.3070417.

Zhang, J., Fan, H., He, D., and Chen, J., 2019. Integrating precipitation zoning with random forest regression for the spatial downscaling of satellite-based precipitation: A case study of the Lancang–Mekong River basin, Int. J. Climatol. 39(10), 3947-3961, https://doi.org/10.1002/joc.6050.

Zhang, T., Li, B., Yuan, Y., Gao, X., Sun, Q., Xu, L., and Jiang, Y., 2018a. Spatial downscaling of TRMM precipitation data considering the impacts of macro-geographical factors and local elevation in the Three-River Headwaters Region, Remote Sens. Environ. 215, 109-127, https://doi.org/10.1016/j.rse.2018.06.004.

Zhang, Y., Tian, Y., Kong, Y., Zhong, B., and Fu, Y., Year. Residual Dense Network for Image Super-Resolution, Proceedings of the IEEE Computer Society Conference on Computer Vision and Pattern Recognition, 2472-2481, https://doi.org/10.1109/CVPR.2018.00262.

Zhao, N., 2021. An efficient downscaling scheme for high-resolution precipitation estimates over a high mountainous watershed, Remote Sensing 13(2), 1-18, https://doi.org/10.3390/rs13020234.